\documentclass[runningheads]{llncs}
\usepackage{graphicx}
\usepackage{amsmath,amssymb} 
\usepackage{color}
\usepackage[width=122mm,left=12mm,paperwidth=146mm,height=193mm,top=12mm,paperheight=217mm]{geometry}
\usepackage{subfig}

\usepackage{cite}
\usepackage{xspace}
\newcommand{\Y}{\mathcal{Y}}

\newcommand*{\ie}{\emph{i.e.}\@\xspace}
\newcommand*{\etal}{\emph{et al.}\@\xspace}
\newcommand*{\etc}{\emph{etc.}\@\xspace}

\def\argmax{\mathop{\rm argmax}	}

\begin{document}
\pagestyle{headings}
\mainmatter

\title{Global Semantic Consistency for \\ Zero-Shot Learning} 


\authorrunning{Fan Wu,~Kai Tian,~Jihong Guan,~Shuigeng Zhou}

\author{Fan Wu$^1$,~Kai Tian$^1$,~Jihong Guan$^2$,~Shuigeng Zhou$^1$\thanks{ Shuigeng Zhou is the corresponding author}}
\institute{$^1$Shanghai Key Lab of Intelligent Information Processing, and School of Computer Science, Fudan University, Shanghai 200433, China,~$^2$Department of Computer Science and Technology, Tongji University, Shanghai 201804, China\\ { \{fanwu15,ktian14,sgzhou\}@fudan.edu.cn,~jhguan@tongji.edu.cn}}

\maketitle

\begin{abstract}

In image recognition, there are many cases where training samples cannot cover all target classes. Zero-shot learning (ZSL) utilizes the class semantic information to classify samples of the unseen categories that have no corresponding samples contained in the training set. In this paper, we propose an end-to-end framework, called Global Semantic Consistency Network (GSC-Net for short), which makes complete use of the semantic information of both seen and unseen classes, to support effective zero-shot learning. We also adopt a soft label embedding loss to further exploit the semantic relationships among classes. To adapt GSC-Net to a more practical setting --- Generalized Zero-shot Learning (GZSL), we introduce a parametric novelty detection mechanism. Our approach achieves the state-of-the-art performance on both ZSL and GZSL tasks over three visual attribute datasets, which validates the effectiveness and advantage of the proposed framework.

\end{abstract}

\section{Introduction}
In some real applications, labeled training samples can not cover all target classes, such as species classification~\cite{akata2015evaluation}, activity recognition~\cite{cheng2013towards} and anomaly detection~\cite{socher2013zero}. Zero-shot Learning (ZSL) ~\cite{palatucci2009zero, akata2013label, socher2013zero} provides a systematic way to address this type of problems by utilizing the semantic information of all classes. The semantic information, such as annotated attributes~\cite{farhadi2009describing}, label word vectors~\cite{mikolov2013distributed} \etc , can be uniformly encoded in attribute vectors~\cite{reed2016learning, Zhang2016Learning}, also referred to as class embedding or (label) semantic embedding.

ZSL uses the samples of the seen classes for training and tests on the samples of the unseen classes. The bridge connecting them is the semantic embeddings of both seen and unseen classes. The essence of ZSL is to learn the association between the visual features and the class embedding, which is then transferred to the samples of unseen classes\cite{Mohammad2014Convex, zhang2016zero, lampert2014attribute,al2016recovering}.

\begin{figure*}[t]
	\centering
        \includegraphics[width=0.8\textwidth]{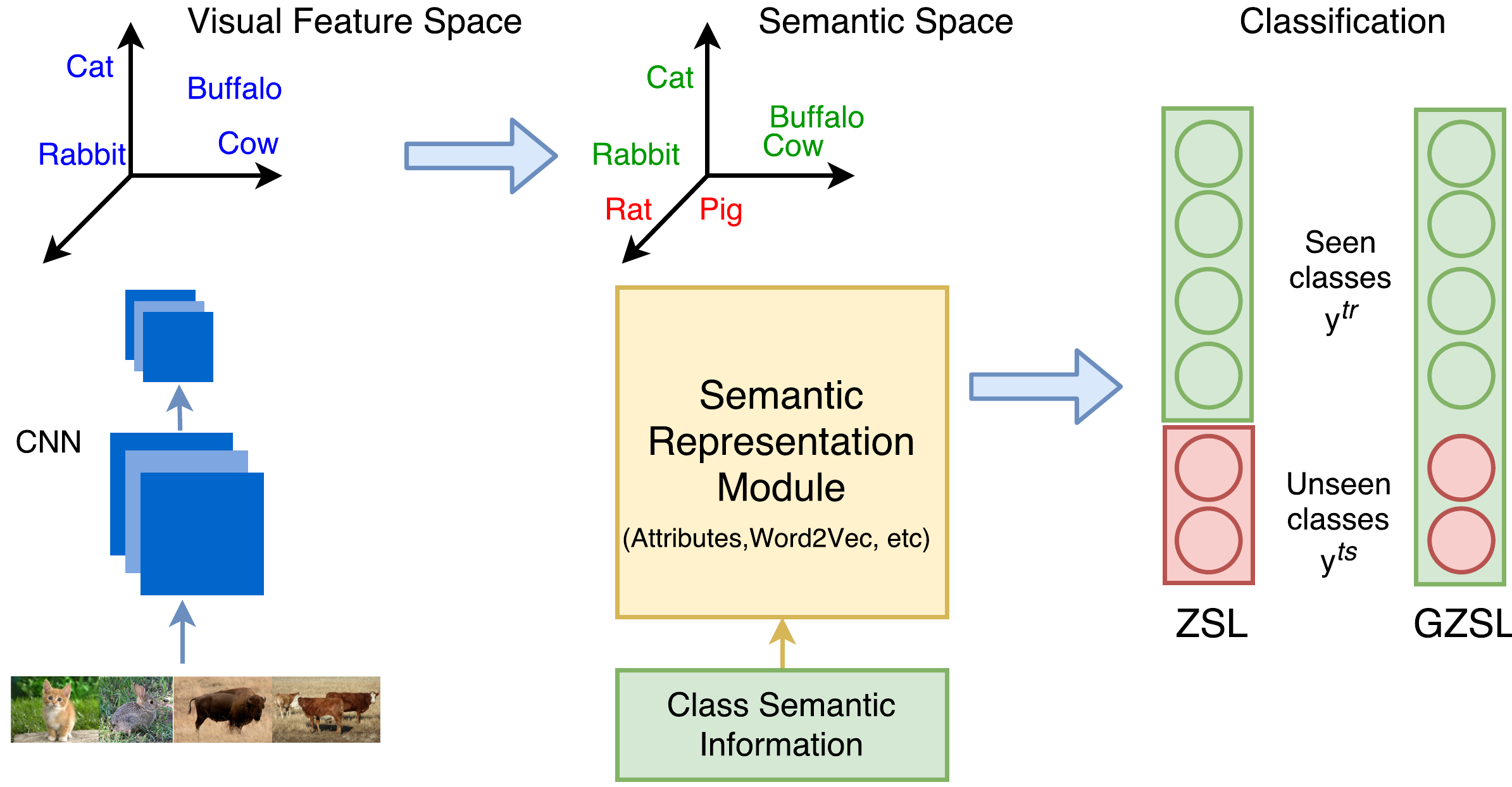}
	\caption{\small Illustration of zero-shot learning (ZSL) and generalized zero-shot learning (GZSL). Available data are labeled images of the seen classes ($\Y^{tr}$) and semantic information of both see and unseen classes ($\Y^{tr+ts}$). In essence, both ZSL and GZSL learn the mapping or compatibilty between visual feature space and semantic space, then apply it to unseen classes. At the test stage, ZSL model is only evaluated on unseen classes ($\Y^{ts}$) whereas GZSL recognizes images from both seen and unseen classes ($\Y^{tr+ts}$).}
	\label{fig:zsl_demo}
\end{figure*}

In the test stage, ZSL considers only classifying new images of unseen classes. However, in some real-world applications, an image classification system usually needs to recognize new images from both seen and unseen classes of the application domain. This is addressed by the so-called \textit{generalized zero-shot learning} (GZSL). Fig.~\ref{fig:zsl_demo} illustrates both zero-shot learning and generalized zero-shot learning tasks.

Most of the existing ZSL methods\cite{xianCVPR17} can be grouped into three types:
\begin{enumerate}
\item These that learn a compatibility function between the image features and the class embeddings, and treat ZSL classification as a compatibility score ranking problem~\cite{frome2013devise, akata2016Label, xian2016latent}. These methods suffer from the following drawbacks: the attribute annotations are pointwise rather than pairwise; compatibility scores are unbounded; and ranking may fail to learn some semantic structure due to the fixed
margin~\cite{annadani2018preserving}.
\item These that project the visual features and semantic embeddings into a shared space and treat ZSL training as ridge regression. The shared space can be visual space, semantic space or a common space between visual and semantic space. The prediction process of these methods is a nearest neighbor search in the shared space, which may cause \textit{hubness problems}~\cite{shigeto2015ridge, radovanovic2010hubs}.
\item In addtion to these methods above, Morgado \etal~\cite{morgado2017semantically} adopted a semantically consistent regularization of the last fully-connected (FC) weight of the neural network in end-to-end training, based on the attribute matrix of the seen classes. However, they did not take the following facts into account: a) different attributes may have different discriminative ability; b) there may be correlation between global class attributes and training sample features.
\end{enumerate}

To overcome the limitations of existing ZSL methods, in this paper we propose an end-to-end framework, called \emph{global semantic consistency network} (GSC-Net) to exploit the semantic embeddings of both seen and unseen classes while preserving the global semantic consistency.
By seeing the global semantic consistency layer as a fully-connected (FC) layer with a fixed weight, we can easily employ all kinds of CNN techniques such as the dropout policy, sigmoid activation, and cross entropy loss. The softmax layer and loss layer in GSC-Net are both over all classes of the learning problem domain, which thus makes full use of the semantic information in training.

Our main contributions are as follows:
\begin{enumerate}
	\item We integrate the global semantic consistency regularization and a neural weighted unit into an end-to-end trainable network.
	\item We adopt the label embedding loss to further exploit the semantic relationships among classes, which thus promotes the neural network to propagate knowledge to the unseen classes.
	\item We introduce a parametric novelty detection mechanism to distinguish between seen and unseen classes for better adapting to the GZSL task.
	\item We validate the effectiveness and advantages of the proposed method by extensive experiments on several popularly used datasets for both ZSL and GZSL tasks.
\end{enumerate}

The rest of the paper is organized as follows: Section~\ref{sec:related} reviews the related works. Section~\ref{sec:method} introduces the proposed approach in detail. The experimental results and analysis are given in Section~\ref{sec:experiments}. The last two sections are discussion and conclusion respectively.

\section{Related Work}
\label{sec:related}
Here, we present a brief review on the related work from four aspects: class semantic embedding, learning bilinear compatibility function, deep ZSL models, and semantically consistent regularization.

\textbf{Class semantic embedding.}
There are several sources of class semantic information: 1) class attribute annotations. They can be discrete or continuous, numerical or categorical. 2) Label word embedding like Word2Vec~\cite{mikolov2013distributed} and glove~\cite{pennington2014glove}. 3) Class hierarchies in Wordnet~\cite{miller1995wordnet}.
Attributes were introduced in \cite{farhadi2009describing, lampert2009learning} and widely used in many following works~\cite{rohrbach2011evaluating, akata2013label, jayaraman2014zero, huang2015learning, kodirov2015unsupervised, changpinyo2016synthesized, xian2016latent}. \cite{akata2013label, akata2015evaluation, rohrbach2011evaluating, xian2016latent} extracted semantic concepts from hierarchies/taxonomies. \cite{Zhang2016Learning} shows that attribute annotations achieve higher accuracy than word vectors, it also presents an end-to-end framework to fuse multiple semantic features. In this paper, we focus on semantic attributes.

\textbf{Learning bilinear compatibility function.} Different from traditional works that simply use a dot product between visual feature representation and semantic representation, some works employ a bilinear compatibility function to combine visual embedding and semantic representation~\cite{frome2013devise,akata2015evaluation,akata2016Label}. The compatibility function is learnable, and is flexible for adaptation.
Generally, $F(x,y;\mathbf{M})=\theta(x)^T\mathbf{M}\phi(y)$, where $\theta(x)$ and $\phi(y)$ represent visual features and semantic vectors respectively, and $\mathbf{M}$ is an intermediate matrix to be learned.
In the test stage, existing compatibility learning frameworks select the class that has the highest compatibility score with the given image.
In the training stage, they usually adopt different loss functions. ALE \cite{akata2016Label} uses the weighted approximate ranking objective~\cite{usunier2009ranking}. ESZSL \cite{romera2015embarrassingly} adopts a square loss to the ranking formulation and
adds an implicit regularization term to the unregularized risk minimization formulation. LATEM \cite{xian2016latent} learns a combination of multiple $\mathbf{M}$s.

\textbf{Deep ZSL models.} Rather than simply taking CNNs as feature extractors, it is more effective to adopt them into the ZSL models, which would provide task specific representations and thus improve the classification performance. Existing end-to-end ZSL models fall into three types. One type maps the visual feature vector to a semantic space by a hinge ranking loss or least square loss~\cite{frome2013devise, socher2013zero}. Another type fuses visual space and semantic space to a common representational space, and then a hinge ranking loss or a binary cross entropy loss is used as the objective function~\cite{yang2014unified, lazaridou2015hubness}. At inference time, those approaches perform nearest neighbor search in a high dimensional space, thus would induce the hubness problem, which is caused by the presence of universal neighbors. In order to deal with such a problem, the third type~\cite{Zhang2016Learning} proposes to map semantic space to visual space, and links visual embedding and semantic embedding with a least square loss.

\textbf{Semantically consistent regularization.}
Morgado \etal~\cite{morgado2017semantically} introduced semantically consistent regularizer to zero-shot recognition. They leveraged the advantages of both independent semantic prediction and semantic embeddings. The attribute codeword regularization is
\begin{equation}
\label{eq:reg}
\Omega[\mathbf{W}] = \parallel \mathbf{W} - \Phi\parallel^2
\end{equation}
where $\mathbf{W}$ is learnable parameter and $\Phi$ is the semantic attribute matrix of seen and unseen classes. By setting $\mathbf{W}$ to be learnable, the attribute space is constrained and consequently performance is improved.

\section{Method}
\label{sec:method}
\subsection{Problem Formulation}
Assume there are $n_{tr}$ seen classes (denoted by a set $\mathcal{Y}^{tr}$) and $n_{ts}$ unseen classes (denoted by a set $\mathcal{Y}^{ts}$) in a problem domain, where seen classes and unseen classes are disjoint, \ie, $\mathcal{Y}^{tr} \cap \mathcal{Y}^{ts} = \varnothing$. So the number of total classes $n_{c} = n_{tr} + n_{ts}$. In the seen class space $\mathcal{Y}^{tr}$, given a dataset with $N_{tr}$ labeled samples, $\mathcal{D}_{tr} = \{(\mathbf{I}_{i}, ~y_{i}), i = 1, \dots , N_{tr}\}$ where $\mathbf{I}_{i}$ means the $i$-th training image, and $y_{i} \in \mathcal{Y}_{tr}$ is the label for $\mathbf{I}_{i}$.

Given the class attribute matrix $\mathbf{W}=[\mathbf{W}^{tr}, \mathbf{W}^{ts}]$ where $\mathbf{W}^{tr} \in \mathbb{R}^{L \times n_{tr}}$ corresponds to the seen classes, $\mathbf{W}^{ts} \in \mathbb{R}^{L \times n_{ts}}$ corresponds to the unseen classes, $L$ is the attribute dimension.

For a new test image $\mathbf{I}_{j}$, the goal of ZSL is to predict the label $\hat{y}_{j}$ just among the unseen classes, \ie, $\hat{y}_{j} \in \mathcal{Y}^{tr}$, while the goal of GZSL is to predict the label $\hat{y}_{j}$ among all classes, \ie, $\hat{y}_{j} \in \mathcal{Y}^{tr+ts}$ where $ \mathcal{Y}^{tr+ts}$ denotes the union space of $\mathcal{Y}^{tr}$ and $\mathcal{Y}^{ts}$ .

\subsection{Global Semantic Consistency Network (GSC-Net) for ZSL}
\label{subsec:GSC_Net}
Morgado \etal~\cite{morgado2017semantically} added a semantically consistent regularization to a CNN network, but they used only the semantic information of seen classes in training. To exploit the semantic attributes of both seen and unseen classes for training, we propose an end-to-end framework, called \emph{Global Semantic Consistency Network} (GSC-Net for short) for the ZSL task.

\begin{figure}[tb]
	\centering
	\includegraphics[width=1.0\textwidth]{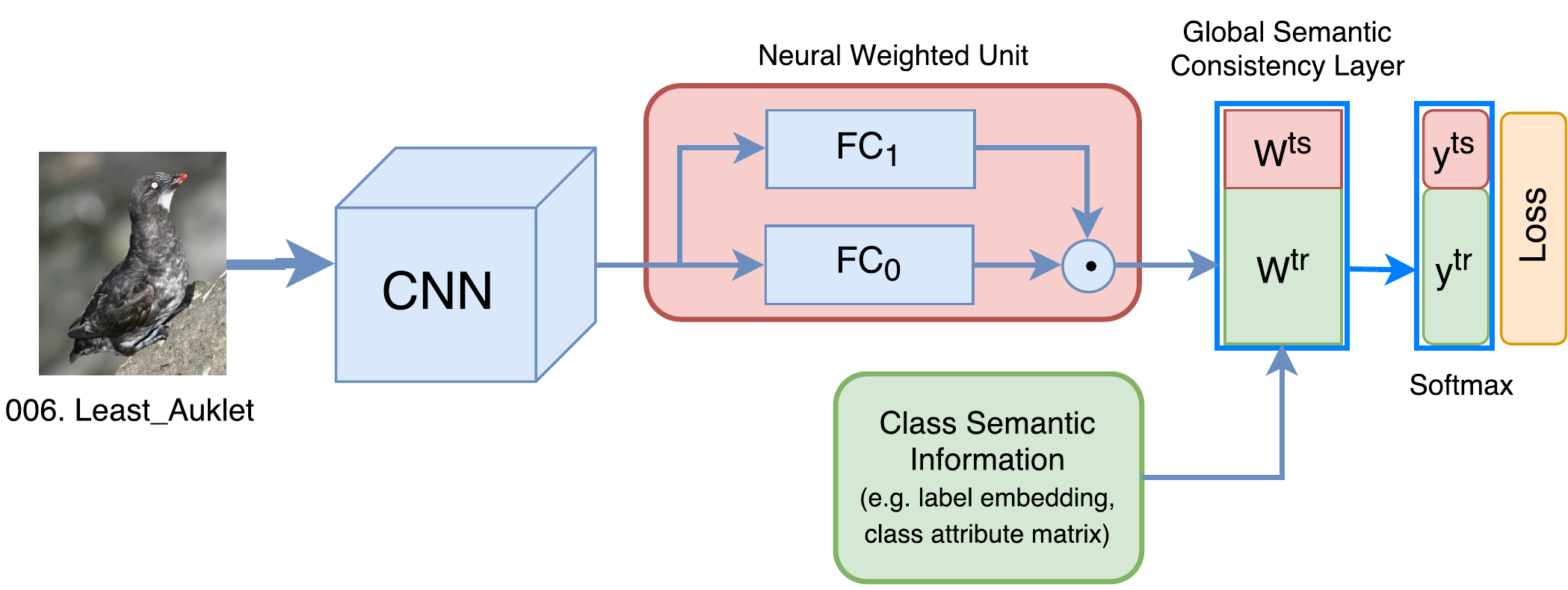}
	\caption{The GSC-Net framework. The class attribute matrix $\mathbf{W}=[\mathbf{W}^{tr}, \mathbf{W}^{ts}]$ where $\mathbf{W}^{tr}$ for the seen/training classes and $\mathbf{W}^{ts}$ for the unseen/test classes. Though no training images belong to the unseen classes, the Global Semantic Consistency (GSC) Layer, softmax layer and loss layer are designed for all classes $\mathcal{Y}^{tr+ts}$.}
	\label{fig:gsc_net}
\end{figure}

As shown in Fig.~\ref{fig:gsc_net}, GSC-Net has four major components as follows:

\begin{enumerate}
	\item \textbf{CNN block}: $x = CNN(\mathbf{I})$. In this paper, we use the pretrained resnet50\cite{He2016Deep} as the CNN by default. The pretrained CNN acts as a feature extractor, with the original last fully-connected (FC) layer being dropped. To make fast end-to-end training, we freeze this block's parameters in the first 5 epochs.
	
    \item \textbf{Neural Weighted Unit}: As shown in Fig.~\ref{fig:gsc_net}, the \textbf{$FC_0$ layer}, \textbf{$FC_1$ layer} and their element-wise product constitute a neural weighted unit (NeWUnit). Specifically,
    1) \textbf{$FC_0$ layer}: $x_0 = W_0x + b_0$. This FC layer maps the CNN features into $L$-dimensional space. Its output can be interpreted as the image embedding.
    2) \textbf{$FC_1$ layer}: $x_1 = \sigma(W_1x + b_1)$. The $FC_1$ layer has a dropout unit and a sigmoid activation. Dropout is a strong regularization and sigmoid squeezes the output to the range [0, 1] as a nonlinear part. In such a way, the training of $FC_0$ and $FC_1$ is asymmetric, which makes the stacked structure ($FC_0$ and $FC_1$) definitely effective. Since the GSC Layer weight $\mathbf{W}^{tr+ts}$ is fixed, $x_1$ can be seen as an adaptive weight for different attributes. The dropout unit can force the model to correctly classify even by using only a part of the attributes, which makes the training process more effective, especially for samples that visually contain only a part of its category's attributes.
    3) \textbf{Element-wise product}: $x_a = x_1 \odot x_0$. We can view this layer's output as the image representation or matching score on each attribute dimension.
	\item \textbf{Global Semantic Consistency (GSC) Layer}: $y^{out} = \mathbf{W}^{tr+ts} \cdot x_a$, where $\mathbf{W}^{tr+ts}$ stands for the global semantic consistency. It can be class attribute matrix, label word embeddings or their combination encoding \etc \cite{Zhang2016Learning} discussed how to fuse multiple semantic vectors together. If the auxiliary information needs a neural encoding layer \cite{Zhang2016Learning}, then we can include this layer in end-to-end co-training. Since most of time the semantic information is about classes and can be fixed for different samples, like the class attribute matrix, we can freeze it in the net, which thus makes it equivalent to a fully connected network with no bias.
	\item \textbf{Loss layer}: Normalize the output score vector to [0, 1] with a softmax $\hat{y} = softmax(y^{out})$. Then, we use cross entropy:
	\begin{equation}
	\label{eq:ce_loss}
		q(\hat{y}, y^{true}) = -y^{true} \cdot log(\hat{y})
	\end{equation}
	where $y^{true}$ is the corresponding one-hot label. Here we do not use weighted approximate ranking loss~\cite{akata2013label} because the class semantic matrix used in experiments is point-wisely labeled.
\end{enumerate}

In this framework, the prediction process can be almost the same in both training stage and test stage by just maximizing the score on classes:
\begin{equation}
\label{eq:pred}
c = \argmax_{j} \hat{y}_j
\end{equation}

\subsubsection{Semantic Consistency \emph{vs}. Global Semantic Consistency.} In order to investigate whether Global Semantic Consistency can give a better supervision on both seen and unseen classes, we also design a Semantic Consistency Network (SC-Net) for comparison experiments. In SC-Net, $\mathbf{W}^{tr}$ and $\mathbf{W}^{ts}$ are respectively used in training stage and test stage, which means the semantic manifold formed by seen classes ($\mathbf{W}^{tr}$) is not aware of the unseen class information ($\mathbf{W}^{ts}$). If we add unseen class information ($\mathbf{W}^{ts}$) in training stage, although unseen class images will not be input to the net, we can still use the global softmax training to form a more comprehensive discriminant space. Intuitively, this can improve performance not only on ZSL tasks, but also on GZSL tasks that recognize both training and test classes ($\Y^{tr+ts}$) at the same time.

In GSC-Net, the softmax and cross entropy loss are also applied to $(n_{tr} + n_{ts})$-dimension output vector $\hat{y}$. Therefore, GSC-Net pays more attention to the attributes mainly owned by unseen classes, which can make the learned features more discriminative among the unseen classes.

\subsubsection{Label embedding loss and soft training.}
With GSC-Net, less seen class images will be misclassified into unseen classes in GZSL, but more unseen class images will be classified into seen classes. This is because the training samples all fall into seen classes $y^{tr}$, making the weights corresponding to $y^{tr}$ larger and larger than those corresponding to $y^{ts}$ during training process.

As the one-hot supervision will cause the net to `lazily' learn a smaller weight for these attributes on which unseen classes have high scores (in the class attribute matrix), so we propose a \textbf{Soft Label Embedding Loss}~(SLE-Loss) by adding a soft label guide to the original cross entropy loss as in \cite{sun2017label}:
\begin{equation}
\label{eq:softLE_loss}
L_{SLE}(\hat{y}, y^{true}, y^{true}_{emb}) = \alpha \cdot q(\hat{y}, y^{true}) + (1-\alpha) \cdot q(\hat{y}, y^{true}_{emb})
\end{equation}
where $q(\cdot)$ is the cross entropy in Eq.~(\ref{eq:ce_loss}), $\hat{y}$ is the output vector of the net, $y^{true}$ is the one-hot vector of the target label while $y^{true}_{emb}$ is the embedding vector of the target label. They are all $n_c$-dimensional. $\alpha$ is a hyperparameter in [0, 1]. Large $\alpha$ will degenerate the loss to standard cross entropy. It is often set around 0.5 if no prior knowledge.

The first part of Eq.~(\ref{eq:softLE_loss}) is the standard one-hot target cross entropy loss $L_{ce}$ while the second part is the soft target cross entropy loss $L_{soft}$.
Since the training samples all belong to seen classes, $L_{soft}$ puts more positive supervision to unseen class attributes.

If $\alpha=0$ then $L_{SLE} = L_{soft}$. In this case, since the FC layers are randomly initialized at the beginning, the projection on each class is almost the same, so the purely soft loss will make the learning process slow at the starting stage. By increasing the value of $\alpha$, we can make training faster and get higher accuracy for seen classes.

We have to utilize the semantic information again to generate the soft label embedding $y^{true}_{emb}$ for all classes $\Y^{tr+ts}$. Inspired by label propagation, we use the class attribute matrix $W$ to build a label graph. Here, we use the adaptive scale policy~\cite{zelnik2005self} to compute the class similarity.
The similarity (or \textbf{affinity}) between two classes is computed by
\begin{equation}
\label{eq:gs_sim}
A_{ij}=\begin{cases}
    e^{-\beta \frac{||W_i - W_j||^2}{h(W_i)h(W_j)}}, & {W_j \in \mathcal{N}(W_i)};\\
    0, & \text{otherwise}.
\end{cases}
\end{equation}
$\mathcal{N}(W_i)$ is the neighbor set of $W_i$, which can be evaluated by setting a distance threshold to reduce the computaion cost. We can also directly replace the relative small $A_{ij}$ values with 0. The \textbf{local scale function} $h(x)$ is defined as
\begin{equation}\label{eq:h_define}
h(x) = ||x-x^{(k)}||
\end{equation}
where $x^{(k)}$ is the $k$-th nearest neighbor of the point $x$.

$\beta$ in Eq.~(\ref{eq:gs_sim}) is a hyperparameter that is used to control the centralization degree of $A$. The larger $\beta$ is, the farther a node is from its neighbor, thus degenerating to the naive one-hot label. Since the local scale function $h(x)$ actually normalizes the numerator term of Eq.~(\ref{eq:gs_sim}), $\beta$ is usually set in the range [1.2, 1.8].

Normalizing $A$ by row, and we get the normalized class embedding matrix $y_{emb} \in \mathbb{R}^{n_c \times L}$.

Overall, SLE-Loss can be applied to many problems with unbalancing training data. We abbreviate the GSC-Net with SLE-Loss as GSC-Net-SLE.

\subsubsection{Relationship to bilinear compatibility methods.}
Our basic architecture can be explained by the classical ZSL methods that use bilinear compatibility function to associate visual and class semantic information, instead of dot product. These methods include Deep Visual Semantic Embedding (DEVISE)~\cite{frome2013devise}, Structured Joint Embedding (SJE)~\cite{akata2015evaluation} and Attribute Label Embedding (ALE)~\cite{akata2016Label}.
Formally, the bilinear compatibility function can be formulated as follows:
\begin{equation}
\label{eq:bilinearCompat}
F(x,y;\mathbf{M}) = \theta(x)^T \mathbf{M} \phi(y)
\end{equation}

\noindent where $\theta(x)$ and $\phi(y)$, \ie, image and class embeddings are computed from image features and class semantic information. $F(.)$ is paramterized by the mapping $\mathbf{M}$ that is to be learned. $x$ is the bottleneck feature of the CNN pretrained on ImageNet2012-1k dataset~\cite{paszke2017automatic}.

On the other side, GSC-Net can be written as
\begin{equation}
\label{eq:lasco}
\hat{y} = \mathbf{W} \cdot (\sigma(W_1x + b_1) \odot x_0)
\end{equation}
where  $\mathbf{W}$, $\sigma(W_1x + b_1)$ and $x_0$ respectively correspond to $\phi(y)$, $\mathbf{M}$ and $\theta(x)$ in Eq.~(\ref{eq:bilinearCompat}). Therefore, the Neural Weighted Unit effectively implements a bilinear compatibility function in GSC-Net.

\subsubsection{Relationship to existing deep ZSL models.}
Many methods \cite{lampert2014attribute,akata2013label,Zhang2016Learning} map the visual features and the label semantic vectors into a shared space, then do classification by computing the nearest label embedding vector:
\begin{equation}
\label{eq:mse}
c = \arg\min_{c} ||\theta(x) - \mathbf{W}_y^{c}||^2
\end{equation}
where $\mathbf{W}_y^{c}$ is the embedding vector of the $c$-th class.
This nearest search method can be clearly visualized and easy to interpret. However, the mean square error is less effective than cross entropy loss in end-to-end training.
So we actually transform the search into a softmax classification. Since $\phi(\mathbf{I}_{j})$ is independent of classification, Eq.~(\ref{eq:mse}) can be written as
\begin{equation}
\label{eq:mse2}
c = \arg\min_{c} - \theta(x)^{T} \mathbf{W}_y^{c} + \frac{1}{2}||\mathbf{W}_y^{c}||^2.
\end{equation}
Since $\mathbf{W}_y^{c}$ is set statistically equal for each class, Eq.~(\ref{eq:mse2}) can be simplified to
\begin{equation}
\label{eq:w2fc}
c = \arg\max_{c} \theta(x)^{T} \mathbf{W}_y^{c}
\end{equation}
where $\theta(x)^{T} \mathbf{W}_y^{c}$ can be seen as expression score on class $c$.
Eq.~(\ref{eq:w2fc}) is equivalent to the last FC layer with no bias in GSC-Net. This maximization process can be integrated into a softmax layer and trained with cross entropy loss.

\subsection{Parametric Novelty Detection for GZSL}
Section \ref{subsec:GSC_Net} introduces our deep ZSL framework GSC-Net. Here we adapt our model for the generalized zero-shot learning (GZSL) task by adding a parametric novelty detection (PND) mechanism.
In GSC-Net-SLE, unseen class images still have relatively high scores on seen classes, which means in most cases $y^{Seen} > y^{Unseen}$ in the output vector.
Therefore, we set a hyperparameter $\gamma$ to control the novelty detection similar to \cite{chao2016empirical}. When
\begin{equation}
\max_{i} y^{Seen}_{i} < \gamma \cdot \max_{j} y^{Unseen}_{j},
\label{eq:gamma_detect}
\end{equation}
we say an unseen class image detected, and take the maximum $y^{Unseen}$ term as the predicted class. So the prediction method with controllable novelty detection goes as follows:
\begin{equation}
\label{eq:gamma_pred}
c =\begin{cases}
    \argmax_{i} y^{Seen}_{i}, & {\max_{i}y^{Seen}_{i} \geq \gamma \cdot (\max_{j} y^{Unseen}_{j})};\\
    \argmax_{j} y^{Unseen}_{j}, & \text{otherwise}.
\end{cases}
\end{equation}
In experiments, $\gamma$ must be larger than 1. The larger the $\gamma$ value, the higher the accuracy on unseen classes.
Our PND mechanism can be easily applied to a typical deep ZSL model. When applied to a certain method, we just add `-PND' to the method's name for notation.

\section{Experiments}
\label{sec:experiments}
Here we present the performance evaluation of the proposed method. We first introduce the three datasets used in experiments and the experimental settings, then give the empirical results on two different tasks: Zero-Shot-Learning (ZSL) and Generalized Zero-Shot Learning (GZSL). 
Especially, we evaluate the contributions of different components of our model to classification performance, and conduct extensive performance comparisons with the existing methods.

\subsection{Datasets and Experimental Settings}
\textbf{Datasets}: Xian \etal \cite{xian2017zero} gave a comprehensive evaluation on the existing ZSL methods on several widely used datasets, and proposed an adapted dataset Animals with Attributes 2 (\textbf{AwA2}) as well as some suggestions on dataset splits for these ZSL datasets. Since our target is to develop a unified end-to-end ZSL framework, we choose 3 datatsets that have open original images and class attribute annotations: \textbf{AwA2}\cite{xian2017zero}, CUB-200-2011 (\textbf{CUB})~\cite{WahCUB_200_2011} and Scene UNderstanding (\textbf{SUN})~\cite{patterson2012sun}. Table~\ref{data_stats} shows more details about them.

In order to make our approach more practical and applicable to more scenarios, we utilize only the class attribute annotations rather than single sample attributes.
It is common in the datasets that the numbers of images in some classes are much larger than in other classes. Therefore, we use the average per-class accuracy to present our results.

\begin{table*}[t!]
\centering
\footnotesize 	
\caption{Details of the ZSL datasets with the proposed splits~\cite{xian2017zero}}
\begin{tabular}{|c|c|c|c|c|c|c|c|}
	\hline
	\textbf{Dataset} & \begin{tabular}[c]{@{}c@{}}No. of \\ attributes\end{tabular} & \begin{tabular}[c]{@{}c@{}}No. of \\ Seen\\ Classes\end{tabular} & \begin{tabular}[c]{@{}c@{}}No. of \\ Unseen\\ Classes\end{tabular} & \begin{tabular}[c]{@{}c@{}}No. of \\ samples\end{tabular} & \begin{tabular}[c]{@{}c@{}}No.of \\ samples\\ (Train)\end{tabular} & \begin{tabular}[c]{@{}c@{}}No. of\\ samples from\\ unseen classes\\ (Test)\end{tabular} & \begin{tabular}[c]{@{}c@{}}No. of\\ samples from\\ seen classes\\ (Test)\end{tabular} \\ \hline
	\textbf{SUN}~\cite{patterson2012sun}   & 102  & 645   & 72  & 14340  & 10320   & 1440   & 2580   \\ \hline
	\textbf{AWA2}~\cite{xian2017zero}    & 85  & 40    & 10      & 37322    & 23527   & 7913    & 5882  \\ \hline
	\textbf{CUB}~\cite{WahCUB_200_2011}     & 312 & 150   & 50      & 11788  & 7057    & 2967  & 1764    \\ \hline 		
\end{tabular}
\label{data_stats}
\end{table*}

\textbf{Settings}: The 2-stage methods use the 2048-D Resnet101\cite{He2016Deep} features provided by \cite{xian2017zero} for all the datasets. To show that our framework can get better results on even smaller CNN base models, we use pretrained Resnet50 \cite{He2016Deep} as our CNN module, which also outputs 2048-D vectors.
In the beginning epochs, since CNN is well pretrained on ImageNet, we can freeze the CNN parameters and train the FC layers only.

Our model contains 4 hyper-parameters as follows:
\begin{itemize}
	\item Dropout keep-ratio $\eta$. $\eta$ is set to 0.5 by default. In experiments, the best results are achieved when $\eta$ is around 0.4$\sim$0.6.
	\item Soft Label Embedding (SLE) Loss ratio $\alpha$. When SLE is used, $\alpha$ is set to 0.5 if no special mention.
	\item The affinity factor $\beta$. For our 3 datasets, we set $\beta = 1.4$ to avoid tuning parameters with test results. Since we use local function, $\beta \in$  $[1.2, 1.8]$ is suitable enough.
	\item Novelty factor $\gamma$. We set $\gamma$ in [1.0, 2.0] for our experiments. If the number of training samples per class is large, which means the seen classes overwhelm unseen classes, $\gamma$ needs to be large. If $\alpha$ is small, the target label will be soft, then small $\gamma$ is considered.
\end{itemize}

\textbf{Training policy}: We use AdaGrad optimizer~\cite{duchi2011adaptive} with a learning rate $10^{-3}$ and a weight decay $\lambda$  of $5 \times 10^{-3}$. $\lambda$ is set large in most cases because our model is strongly regularized by the semantic layer in Equ.~(\ref{eq:w2fc}). We use pyTorch\cite{paszke2017automatic} and run our experiments on Titan Xp GPUs with early stopping policy.

\subsection{Zero-Shot Learning Experiments}
The results on three datasets are presented in Table~\ref{tb:zsl_results}.
The upper part shows the 2-stage (opposite to end-to-end) methods whose results are reported in \cite{xian2017zero}.  ALE~\cite{akata2016Label} is simple but effective on all datasets. These methods all use 2048-D ResNet101 features. The lower part stands for end-to-end approaches. Under the same protocol, we implement Deep-SCoRe and DEM on resnet50 and test our 3 models on ZSL: SC-Net, GSC-Net, GSC-Net-SLE.

On all 3 datasets, SC-Net outperforms Deep-SCoRe and DEM by an explicit margin, which shows the neural weighted unit performs better due to its nonlinear property. On the basis of SC-Net, GSC-Net improves performance a lot by making full use of the total class attribute matrix and boosting the feature learning for unseen classes. With soft training, GSC-Net-SLE further lifts the performance. Overall, GSC-Net-SLE surpasses the existing methods and achieves the state-of-the-art performance on all 3 datasets.

Comparing the end-to-end (E2E) methods and 2-stage (2S) methods, we can easily discover that E2E methods exceed 2S methods significantly on AWA2 and CUB, but hit a draw on SUN.
The reasons may be: 1) there are only 16 images per seen class in SUN, which does not contribute much to CNN finetuning. 2) There are 717 classes but only 102 attributes annotated in SUN. Note that the dimension of the class attribute matrix $W$, \ie, the last FC weight, is 717$\times$102, therefore the feature dimensionality of 102 is not large enough for 717-way classification.

\begin{table}[t!]
	\centering
	\normalsize
\caption{Average per-class accuracy (top-1 in \%) for ZSL task. Results of the 2-stage approaches are from \cite{xian2017zero}.}
	\label{tb:zsl_results}
	
\begin{tabular}{|c|c|c|c|c}
		\hline
		\textbf{Method}            & \textbf{SUN}  & \textbf{AWA2} & \textbf{CUB} \\ \hline
		DAP~\cite{lampert2014attribute}   & 39.9   & 46.1   & 40.0    \\
		IAP~\cite{lampert2014attribute}  & 19.4   & 35.9  & 24.0   \\
		CONSE~\cite{Mohammad2014Convex}      & 38.8    & 44.5          & 34.3     \\
		CMT~\cite{socher2013zero}        & 39.9          & 37.9          & 34.6   \\
		SSE~\cite{zhang2015zero}        & 51.5          & 61.0          & 43.9    \\
		LATEM~\cite{xian2016latent}      & 55.3          & 55.8          & 49.3     \\
		ALE~\cite{akata2016Label}    & \textbf{58.1} & \textbf{62.5}  & 54.9      \\
		DEVISE~\cite{frome2013devise}     & 56.5          & 59.7    & 52.0    \\
		SJE~\cite{akata2015evaluation}  & 53.7   & 61.9    & 53.9    \\
		ESZSL~\cite{romera2015embarrassingly}  & 54.5     & 58.6  & 53.9    \\
		SYNC~\cite{changpinyo2016synthesized}   & 56.3   & 46.6  & \textbf{55.6}   \\
		SAE~\cite{Kodirov_2017_CVPR}    & 40.3    & 54.1     & 33.3    \\ \hline
	    Deep-SCoRe~\cite{morgado2017semantically}(Resnet50) & 51.7 & 69.5 & 61.0 \\
        DEM~\cite{Zhang2016Learning}(Resnet50) & 51.1 & 68.7 & 60.1 \\
        RELATION NET~\cite{sung2017learning}(GoogleNet)  & - & - & 62.0   \\ \hline
     SC-Net(our benchmark, Resnet50) & 53.6 & 72.9 & 64.9   \\
     GSC-Net(ours, Resnet50) & 57.6 & 74.9 & 68.2   \\
	 GSC-Net-SLE(ours, Resnet50) & \textbf{58.1} & \textbf{75.2} & \textbf{69.2}  \\ \hline
	\end{tabular}
	\end{table}

\subsection{Generalized Zero-shot Learning Experiments}

\begin{figure}[tb]
    \centering
    \subfloat[SUN]{\label{fig:gzsl_sun}{\includegraphics[width=0.46\textwidth]{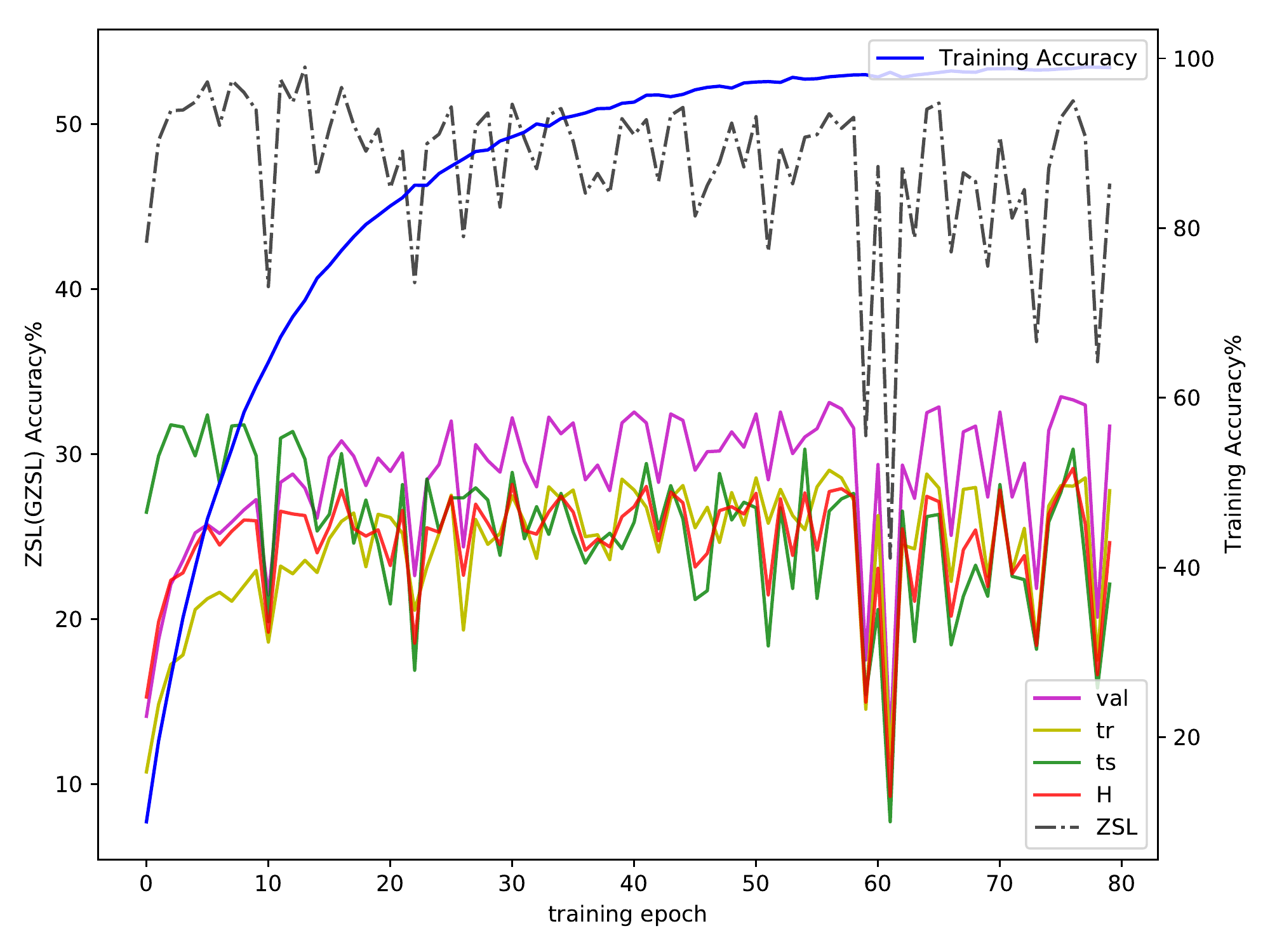} }}
    \qquad
    \subfloat[CUB]{\label{fig:gzsl_cub}{\includegraphics[width=0.46\textwidth]{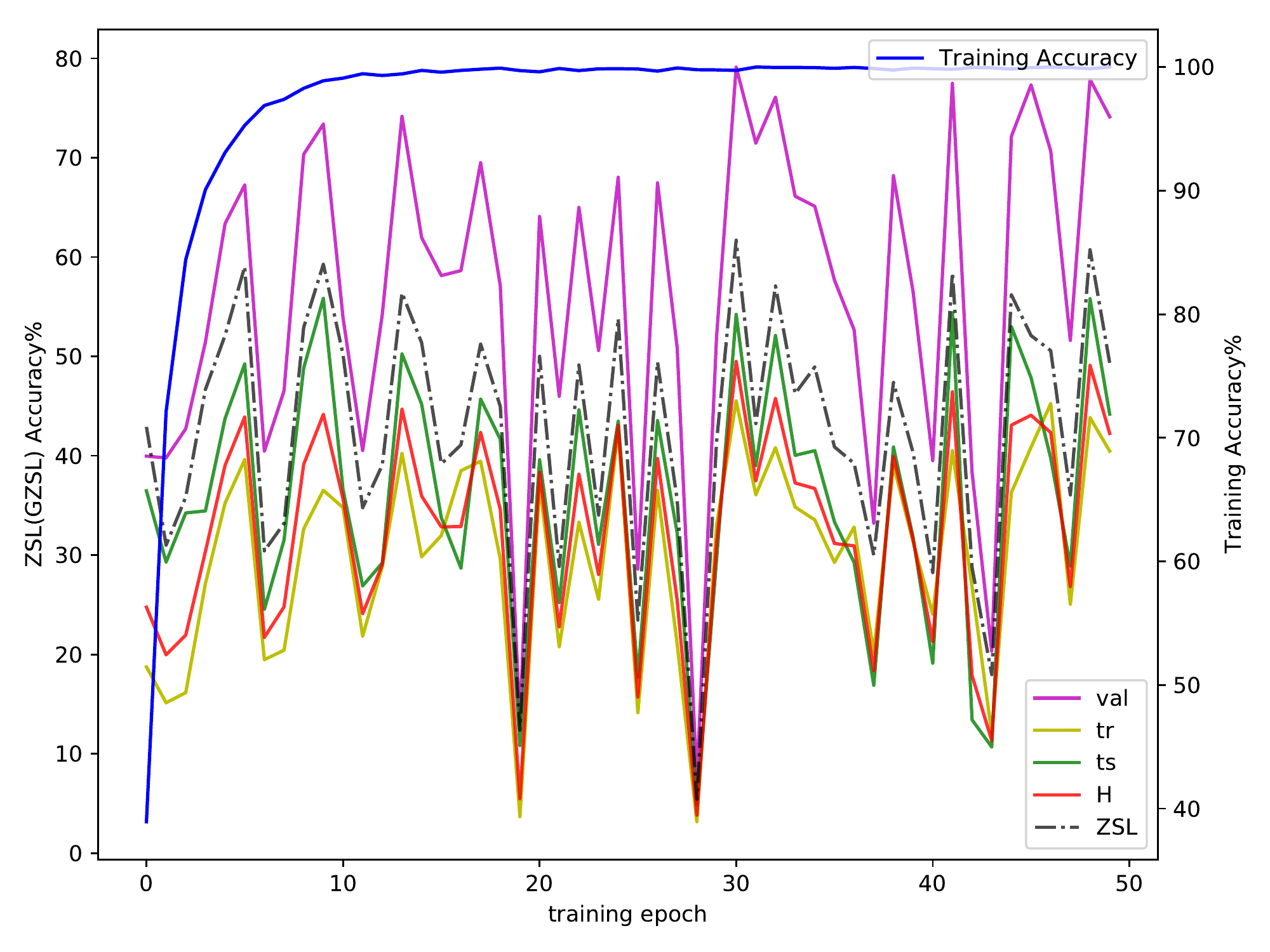} }}
      \caption{\small GSC-Net-SLE-PND~($\alpha=0.5$) training process on SUN\protect\subref{fig:gzsl_sun} and CUB\protect\subref{fig:gzsl_cub} for ZSL task and GZSL task respectively. The blue and purple lines refer to the training and validation accuracy on seen classes. 
      The left axis corresponds to ZSL (GZSL) accuracy while the right axis for training accuracy.}
     \label{fig:gzsl_train}
\end{figure}

\begin{table*}[t!]
	\centering
	\normalsize
\caption{Results on GZSL task. 
The results of the existing approaches are taken from \cite{xian2017zero}. CMT* refers to CMT~\cite{socher2013zero} with novelty detection. `-PND' refers to employing our novelty detection mechanism.}
	\label{gzsl_results}	
\resizebox{\textwidth}{!}{\begin{tabular}{|c|c|c|c|c|c|c|c|c|c|}
		\hline
		& \multicolumn{3}{c|}{\textbf{SUN}}             & \multicolumn{3}{c|}{\textbf{AWA2}}            & \multicolumn{3}{c|}{\textbf{CUB}}           \\ \hline
		\textbf{Method}   & \textbf{ts}            & \textbf{tr}            & \textbf{H}             & \textbf{ts}            & \textbf{tr}            & \textbf{H}             & \textbf{ts}            & \textbf{tr}            & \textbf{H}           \\ \hline
		DAP~\cite{lampert2014attribute}      & 4.2           & 25.1          & 7.2           & 0.0           & 84.7          & 0.0           & 1.7           & 67.9          & 3.3   \\
		IAP~\cite{lampert2014attribute}      & 1.0           & 37.8          & 1.8           & 0.9           & 87.6          & 1.8           & 0.2           & \textbf{72.8} & 0.4   \\
		CONSE~\cite{Mohammad2014Convex}     & 6.8     & 39.9 & 11.6          & 0.5       & \textbf{90.6} & 1.0      & 1.6    & 72.2     & 3.1      \\
		CMT~\cite{socher2013zero}       & 8.1           & 21.8          & 11.8          & 0.5           & 90.0          & 1.0           & 7.2           & 49.8          & 12.6       \\
		CMT*~\cite{socher2013zero}          & 8.7           & 28.0          & 13.3          & 8.7           & 89.0          & 15.9          & 4.7           & 60.1      & 8.7   \\
		SSE~\cite{zhang2015zero}               & 2.1           & 36.4          & 4.0           & 8.1           & 82.5          & 14.8          & 8.5           & 46.9          & 14.4  \\
		LATEM~\cite{xian2016latent}           & 14.7          & 28.8          & 19.5          & 11.5          & 77.3          & 20.0          & 15.2          & 57.3          & 24.0      \\
		ALE~\cite{akata2016Label}   & \textbf{21.8} & 33.1          & \textbf{26.3}          & 14.0          & 81.8          & 23.9          & 23.7          & 62.8      & \textbf{34.4}  \\
		DEVISE~\cite{frome2013devise}        & 16.9          & 27.4          & 20.9          & \textbf{17.1}         & 74.7          & \textbf{27.8}          & \textbf{23.8}          & 53.0          & 32.8     \\
		SJE~\cite{akata2015evaluation}       & 14.7          & 30.5          & 19.8          & 8.0           & 73.9          & 14.4          & 23.5          & 59.2          & 33.6     \\
		ESZSL~\cite{romera2015embarrassingly}      & 11.0          & 27.9          & 15.8          & 5.9           & 77.8          & 11.0          & 12.6          & 63.8          & 21.0     \\
		SYNC~\cite{changpinyo2016synthesized}      & 7.9         & \textbf{43.3}   & 13.4          & 10.0          & 90.5          & 18.0          & 11.5          & 70.9          & 19.8    \\
		SAE~\cite{Kodirov_2017_CVPR}               & 8.8           & 18.0          & 11.8          & 1.1           & 82.2          & 2.2           & 7.8           & 54.0          & 13.6    \\ \hline
	 DeepSCoRe-PND~\cite{morgado2017semantically} & 17.3 & 30.8 & 22.2 & 8.8 & 91.1 & 16.0 & 20.3 & 65.8 & 31.0\\
     SC-Net & 10.3 & 33.4 & 15.8 & 3.8 & \textbf{93.4} & 7.2 & 15.0 & \textbf{70.1} & 24.7 \\
     SC-Net-PND & 26.2 & 27.7 & 26.9 & 10.6 & 92.8 & 19.1 & 30.2 & 59.1 & 39.9\\
     GSC-Net-PND & 26.5 & \textbf{39.1} & 31.6 & 16.6 & 93.2 & 28.1 & 39.9 & 64.7 & 49.3\\
     GSC-Net-SLE-PND & \textbf{29.2} & 35.8 & \textbf{32.2} & \textbf{19.2} & 91.4 & \textbf{31.7} & \textbf{49.9} & 62.4 & \textbf{55.4}  \\  \hline
	\end{tabular}}
	\end{table*}

In GZSL setting, the search space contains both the seen classes and the unseen classes. We use the same evaluation protocol as in \cite{xian2017zero}. Let \textbf{ts} stand for GZSL accuracy on unseen classes and \textbf{tr} for GZSL accuracy on seen classes. \textbf{H} is the harmonic mean between \textbf{ts} and \textbf{tr} as follows:
\begin{equation}
H = \frac{2* tr *ts}{tr + ts}
\end{equation}
\textbf{H} pays attention to the smaller one between \textbf{tr} and \textbf{ts}, it is a balanced evaluation for the GZSL task.

Fig.~\ref{fig:gzsl_train} shows the training process of GSC-Net-SLE-PND ($\alpha=0.5$) on SUN and CUB for ZSL task and GZSL task respectively. We can see that \textbf{ts} for unseen classes in GZSL is much lower than the ZSL accuracy for seen classes, which shows that GZSL is a much harder task than ZSL.

The model reaches a high accuracy in less than 20 epochs and then oscillates irregularly, so we save the earlier models with early stopping policy. Fig.~\ref{fig:gzsl_train} also shows that ZSL/GZSL accuracy fluctuates with the validation accuracy $val$ (purple line in Fig. \ref{fig:gzsl_train}) almost in the same pace. This directly reveals that better feature learning gives better ZSL/GZSL prediction. So we can use the validation accuracy to select the saved models in a real scenario.

Table~\ref{gzsl_results} reports the results of GZSL on the three datasets. In the upper part, we can see that most existing ZSL methods perform very poorly on GZSL task (indicated by \textbf{H}). CMT~\cite{socher2013zero} proposes novelty detection, which improves its performance a lot on GZSL. However, they all have very low accuracy on unseen classes in GZSL, which leaves them quite low on \textbf{H}.

Our approaches get the state-of-the-art results on GZSL task and surpass others by a large margin on all 3 datesets. The reason is threefold:
\begin{enumerate}
	\item GSC-Net-SLE-PND can learn more complex features with the nonlinear unit and more comprehensive features with global semantic consistency.
	\item GSC-Net-SLE-PND uses soft label embedding loss to put more positive supervision on the unseen classes' attributes by using a smaller $\alpha$.
	\item GSC-Net-SLE-PND employs a parametric novelty detection mechanism to control the novelty threshold with $\gamma$.
\end{enumerate}

For the three datasets, GSC-Net-SLE-PND improves performance most significantly on CUB, with \textbf{H} from 34.4\% to 55.4\%, where we actually balance the \textbf{ts} and \textbf{tr} so very well by setting a suitable $\gamma$. For SUN, there are too many classes and only 16 images per training seen class, which makes it a challenging problem to get high accuracy on both ts and tr. Oppositely, AWA2 faces an extremely unbalancing situation: there are so many training images that unseen classes are totally overwhelmed by seen classes.

\subsection{Effectiveness of FC Weight as Class Semantic Embedding}
The good perfermance of our global semantic consistency framework proves that setting the class attribute matrix as the weight $W$ of the last FC layer of CNN can perfectly exploit both seen and unseen class semantic information in training. This implies that $W$ in a fully trained network can be seen as the semantic embedding for the classes.

To verify this guess, we train CNN from scratch with a last no-biased FC layer on several datasets: cifar100~\cite{krizhevsky2009learning}, fashion-mnist~\cite{xiao2017online}, ImageNet-1k~\cite{ILSVRC15}. Then, we use tsne~\cite{maaten2008visualizing} to reduce the dimension of their $W$ to 2D and visualize it. Due to space limit, here we visualize only the results on cifar100 in Fig.~\ref{fig:cifar100_w}. The other results are included in the supplemental materials.

In Fig.~\ref{fig:cifar100_w}, we can see that semantically similar labels are embedded at neighboring positions, such as bicycle and motorcycle, oak\_tree and maple\_tree, leopard and tiger \etc This visualization conforms to our interpretation of the last FC layer.
\begin{figure}[tb]
	\centering
	\includegraphics[width=0.8\textwidth]{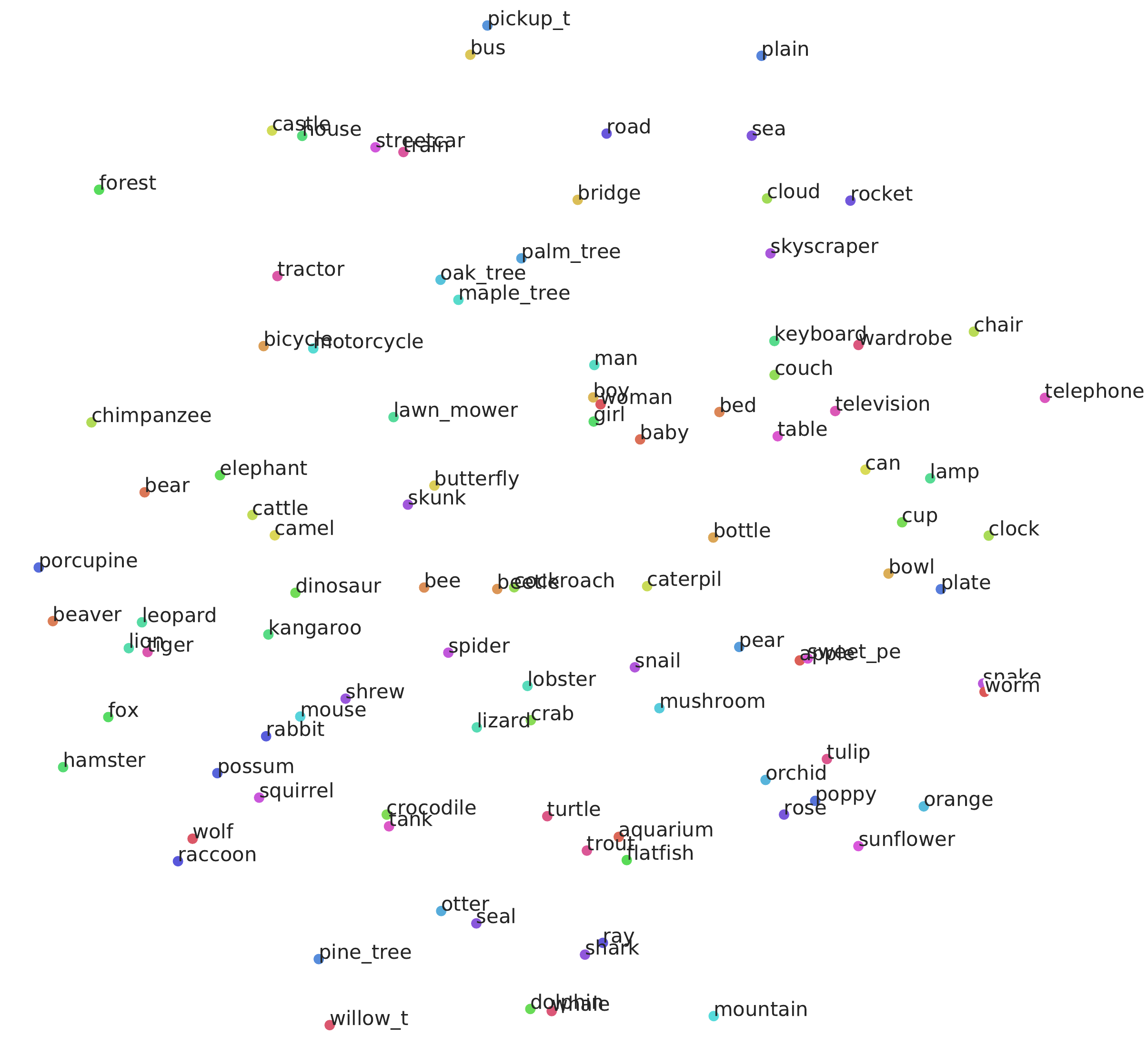}
	\caption{\small TSNE visualization of the last FC weight in a trained CNN on cifar100.}
	\label{fig:cifar100_w}
\end{figure}

\section{Discussion}
Our framework includes unseen classes in softmax at training time. Therefore, the test process in GZSL is exactly the same as the training process, which actually makes our framework consistent with normal supervised neural network. This means when new samples of unseen classes are available, our framework can normally use them in training without changing anything. Therefore, our method can be easily adapted into an online end-to-end learning system. For example, it can be a direct solution for cold start problem in deep collaborative filtering.

Although the perfermance of the proposed approach is outstanding, there are two problems that need to be mentioned. On the one hand, when the number of unseen classes $n_{ts}$ is large, the FC layer weight matrix $W \in \mathbf{R}^{L \times (n_{tr}+n_{ts})} $ of GSC-Net will be too big to efficiently train.
On the other hand, our method relies heavily on the class embedding matrix $\textbf{W}$. For the cases where $\textbf{W}$ is not accurate or very noisy, like word vectors, further investigation is needed.

\section{Conclusion}

In this work, we try to make full use of the global class semantic information to
improve the classification performance in ZSL and GZSL. We propose an end-to-end model with a neural weighted unit to increase the learning ability under global semantic constraints. We also adopt the label embedding loss to further exploit the semantic relationships between classes, which thus enables the neural network to propagate more knowledge to unseen classes. Last but not least, we introduce a simple but effective novelty detection mechanism with a controllable parameter.
Our approaches obtain the state-of-the-art results on three datasets for both ZSL and GZSL tasks. Our experiments also show the effectiveness of FC weight as a class semantic embedding.


\bibliographystyle{splncs}
\bibliography{egbib}
\end{document}